\documentclass[letterpaper, 10 pt, conference]{ieeeconf}  

\IEEEoverridecommandlockouts                              

\usepackage{amsmath,amsfonts}
\usepackage{algorithmic}
\usepackage{algorithm}
\usepackage{array}
\usepackage[caption=false,font=normalsize,labelfont=sf,textfont=sf]{subfig}
\usepackage{textcomp}
\usepackage{stfloats}
\usepackage{url}
\usepackage{verbatim}
\usepackage{ifpdf}
\ifpdf
  \usepackage{graphicx}
  \usepackage{hyperref}
\else
  \usepackage[dvipdfmx]{graphicx}
  \usepackage[dvipdfmx]{hyperref}
\fi
\usepackage{bm}
\usepackage{here}
\usepackage{comment}
\usepackage{cite}
\usepackage{color}

\title{\LARGE \bf
Slip-Induced Fall Suppression During Cornering for a Remotely Operated Motorcycle with a Rear-Wheel-Swing Mechanism
}

\author{Tadashi Sumioka$^{1}$, Kazushi Akimoto$^{1}$,  Takuya Tsujimura$^{1}$, Sho Takayanagi$^{1}$ and Kineo Tomura$^{2}$\\
\thanks{*This work was not supported by any organization}
\thanks{$^{1}$Tadashi Sumioka, Kazushi Akimoto, Takuya Tsujimura, Sho takayanagi and Tomoyo Konya are with Honda R\&D Co., Ltd., Innovative Research Excellence, Frontier Robotics, Saitama 3510114, Japan
        {\tt\small tadashi\_sumioka@jp.honda}
        }%
\thanks{$^{2}$Kineo Tomura is with Honda Motor Co., Ltd. }
}

\begin{document}

\maketitle
\thispagestyle{empty}
\pagestyle{empty}

\begin{abstract}
Slip-induced falls when riding motorcycles are difficult to avoid even for expert riders and represent a major cause of single-vehicle accidents. Despite their prevalence, no practical solution has yet been proposed to suppress this phenomenon. We previously introduced a motorcycle model applicable to systems with both front- and rear- wheel steering. In this paper, we extend that model and formulate the suppression of slip-induced falls as an optimization problem that distributes assistive control inputs according to vehicle speed and road surface conditions. To evaluate the proposed method, we developed a remotely operated motorcycle and conducted experiments under two scenarios: slalom driving on a low-friction surface at 5.5 m/s, and turning maneuvers on synthetic ice to simulate driving over manholes on rainy days. The results demonstrate that our approach successfully suppresses falls under these challenging conditions.
\end{abstract}


\section{Introduction}
\par
Motorcycles are attractive vehicles because they are small, highly maneuverable, and environmentally friendly. 
However, when a motorcycle falls over, it can cause serious injury to the rider. 
A lot of research has thus been conducted on the dynamic performance of motorcycles (e.g., \cite{sharp}, \cite{cossalter}, \cite{tanell}), particularly on the suppression of falls.
\par
The causes of motorcycle falls can be divided into two categories depending on the vehicle  speed. When a motorcycle is stopped or moving at an extremely low speed,
falls are due to the fundamental lack of stability of motorcycles. Many methods have thus been developed to improve stability, such as those that add flywheels or movable axles (e.g.,\cite{singhania}, \cite{rosso}, \cite{akimoto}, \cite{tsujii}, \cite{sumioka}). 
Although there are still issues to be addressed, such as the coordination between the rider and fall suppression control and implementation costs, progress has been made in terms of fall prevention. 
When a motorcycle is moving at a normal speed, falls are often caused by a loss of traction. 
\par
Falls that occur at normal speeds can be further divided into falls that occur while going straight and falls that occur while turning. 
Technologies for suppressing slippage and falls while going straight include an anti-lock
braking system\cite{ABS1}, which prevents the wheels from locking during braking, and a traction control system\cite{TCS}, which prevents the rear wheels from spinning due to an excessive driving force. 
These two technologies, which are used in many mass-produced motorcycles, detect the longitudinal slip ratio of each wheel and control the braking and driving force such that the slip ratio does not become excessive. 
In contrast, there is little research on preventing motorcycle slips and falls while turning. 
A slip and fall avoidance method developed by Bosch \cite{BOSCH} uses gas injection to generate a restoring moment; however, this makes the method difficult to use in mixed traffic situations  involving cars, motorcycles, and pedestrians. 
The present study proposes an extension of a previously developed front- and rear-wheel steering control method \cite{sumioka} to achieve slip-induced fall suppression while preserving the fundamental characteristics of the motorcycle structure. 
\par
Considering rider safety, it is difficult to test a motorcycle in terms of slipping and falling while turning. 
Furthermore, although tire–road interaction modeling has been investigated in works such as \cite{cossalter}, the behavior in limit conditions—e.g., rollover—does not match real vehicle dynamics, making simulation-based verification difficult. 
To overcome the above problems, experiments can be conducted using a remotely operated motorcycle. 
Riderless motorcycles have been investigated mostly using simulations; few studies have verified their effectiveness using actual vehicles (e.g., \cite{auto_motorcycle1} --\cite{auto_motorcycle5}). 
BMW \cite{BMW} and Yamaha \cite{MOTOBOT} have developed actual autonomous motorcycles, but neither has published a related study. 
Miyagishi et al. \cite{miyagishi} used an actual autonomous motorcycle in their study. Their stabilization control method outputs a front-wheel steering angle that makes the roll angle track the target roll angle, as done in self-driving bicycles \cite{auto_bicycle}. However, there are no reports on the behavior when slippage occurs. 

Although autonomous bicycles using gyroscopes—such as those described in \cite{auto_bicycle2}—have been proposed, this approach is not practical for slip‑induced rollover avoidance in motorcycles operating in normal speed ranges. This is because a significantly larger restoring roll moment is required, and the associated actuator output and power demand become impractically high.
\par
\begin{figure}
\centering
\resizebox*{8cm}{!}{\includegraphics{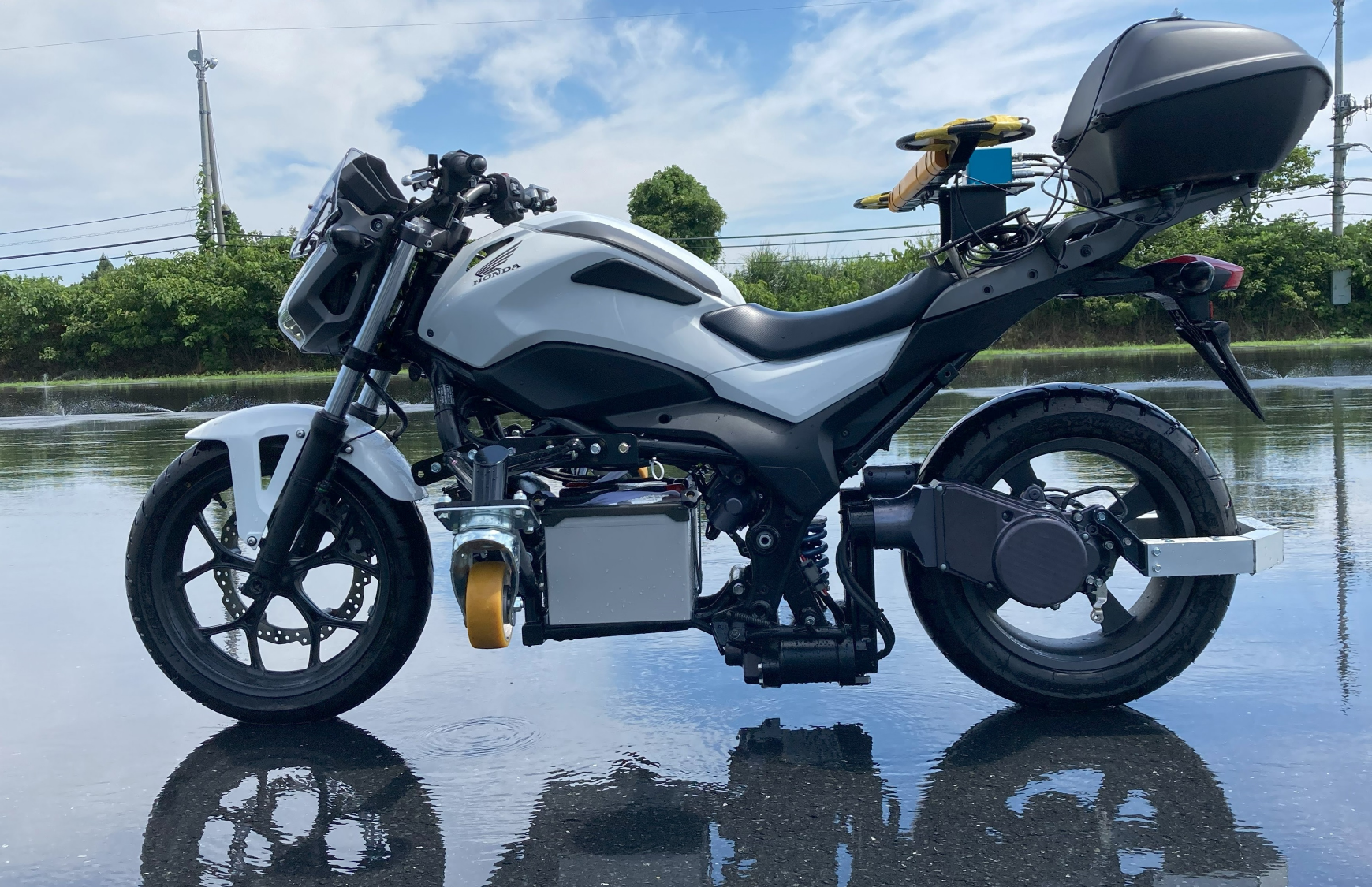}}\hspace{5pt}
\caption{Motorcycle with rear-wheel swing assist mechanism. This experimental vehicle was developed based on a mass-produced motorcycle.} \label{fig:riding_assist_vehicle}
\end{figure}

In this study, we propose a method for suppressing slips and falls while turning and verify it using a remotely operable motorcycle, shown in Fig.~\ref{fig:riding_assist_vehicle}.
Based on a previously developed vehicle \cite{sumioka}, we propose an optimal front- and rear-wheel assist method that takes into account vehicle speed and the road surface friction coefficient for driving speeds that range from extremely low to 5.5 m/s while maintaining the divergent component of motion (DCM) feedback control, which is the best regulator in terms of stabilization.
In order to establish a fall‑prevention method safely, rider-on-board experiments are beyond our present scope. However, when a rider is on board, assist control is expected to require additional rider–machine cooperative control methods, as suggested in \cite{sumioka}. Therefore, future work will extend this approach toward systems capable of interacting with a rider.

\section{PRELIMINARIES}
Here, we briefly explain the equivalent two-point mass decomposition linear inverted pendulum (E2PMD-LIP) model and the DCM feedback control method proposed in \cite{sumioka}.
\subsection{Equivalent two-point mass decomposition linear inverted pendulum model}
\begin{figure*}
\centering
\resizebox*{15cm}{!}{\includegraphics{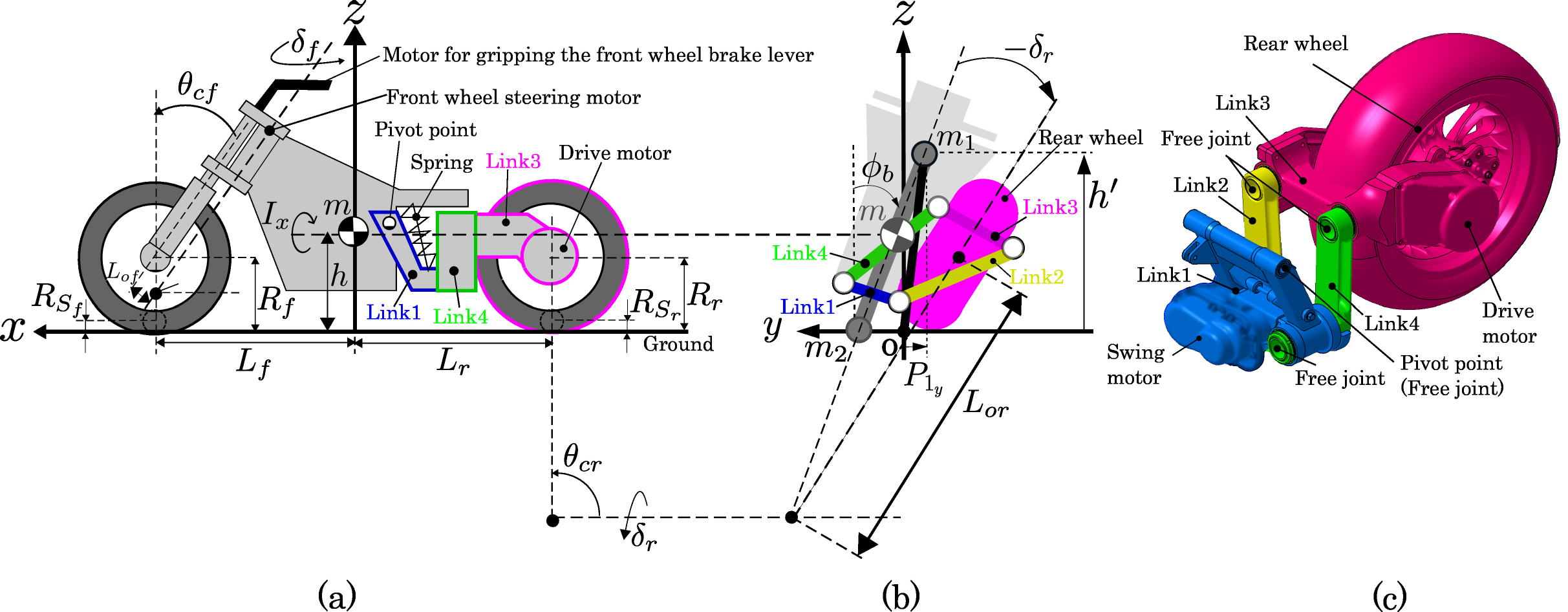}}\hspace{5pt}
\caption{Hardware configuration of motorcycle with rear-wheel swing assist mechanism. (a) Left-side view. (b) Rear view. (c) Details of four-bar linkage mechanism. The front-wheel steering has a motor that can control the angular velocity. The rear wheel is swung using a four-bar linkage mechanism, where the length of Link3 is longer than that of Link1; Link2 and Link4 are the same length. The vehicle body is fixed to Link1 and the rear wheel is fixed to Link3. Link1 is connected to the vehicle body at the pivot point. $\delta_{r}$ is the rear-wheel swing angle, which is the same as the rear-wheel roll angle with respect to the vehicle body. Compared with $\theta_{cf}$ and $L_{of}$, $\theta_{cr}$ and $L_{or}$ are much larger; however, $\delta_{r}$ can still be regarded as the rear‑wheel steering angle, in a broad sense, based on the same definition as that of the front wheel.} \label{fig:hard}
\end{figure*}

The variables used in this study are defined in Table \ref{table:variables}. 
Fig.~\ref{fig:hard} shows the hardware configuration that enables swing of the rear wheel using a four-bar linkage mechanism. 
Here, the $x–y–z$ coordinate system is a frame that translates in the plane together with the vehicle body.
\begin{table*}[hbtp]
  \caption{Definitions of Variables}
  \label{table:variables}
  \centering
\begin{tabular}{lll}
    \hline
    Symbol  & Unit &Description   \\
    \hline \hline
    $m $  & kg  & Total mass \\
    $h $  & m  & Height of center of gravity \\
    $I_{x}$  & kgm$^{2}$  & Total roll inertia about center of gravity \\
    $g $  & m/s$^{2}$  & Gravitational acceleration \\
    $\delta_{i}$  & rad  & Steering angle \\
    $R_{i}$  & m  & Wheel radius \\
    $R_{S_{i}}$  & m  & Wheel sectional radius \\
    $\theta_{ci}$  & rad  & Caster angle \\
    $L $  & m  & Wheel base \\
    $L_{i}$  & m  & Length from wheel axle to center of gravity \\
    $L_{oi}$  & m  & Fork offset \\
    $\phi_{b}$  & rad  & Roll angle of vehicle body (lean angle) \\
    $\beta$  & rad & Side-slip angle of the center of gravity, measured only for slip detection \\
                 &      &and not used in the controller. \\
    \hline
  \end{tabular}
\end{table*}

Using the E2PMD-LIP model proposed in \cite{akimoto} and \cite{sumioka}, the linear inverted pendulum (LIP) point mass $m_{1}$, its height $h'$, and the point mass on the ground $m_{2}$ can be obtained as follows:
\begin{eqnarray}
h' &=& h + \frac{I_{x}}{mh}, \\
m_{1} &=& \frac{mh}{h'}, \\
m_{2} &=& \left( 1  -\frac{h}{h'} \right)m.
\end{eqnarray}
Furthermore, the dynamics of the E2PMD-LIP model can be expressed by the following state equation:
\begin{eqnarray}
\frac{d}{dt}
\left[ 
\begin{array}{c} 
P_{1_{y}} \\ 
\dot{P}_{1_{y}}
\end{array} 
\right ]
=
\left[ 
\begin{array}{cc} 
0 & 1 \\ 
\omega^{2} & 0
\end{array} 
\right ]
\left[ 
\begin{array}{c} 
P_{1_{y}} \\ 
\dot{P}_{1_{y}}
\end{array} 
\right ]
+
\left[ 
\begin{array}{c} 
0 \\ 
-1
\end{array} 
\right ]
u,  \label{state_eq}
\end{eqnarray}
where $P_{1_{y}}$ is the lateral displacement of
the mass point $m_{1}$. 
$\omega$ and $u$ are the natural frequency and lateral acceleration input in the E2PMD-LIP model, respectively; they are expressed as follows: 
\begin{eqnarray}
\omega\! \! &=& \! \! \sqrt{\left( 1 - \frac{m}{m_{1}h'}R_{S_{eq}} \right)\frac{g}{h'}} \label{a_21}, \label{omega_bike}\\
u \! \! &=& \! \!  u_{v} + u_{v_{0}}. \label{u}
\end{eqnarray}
$R_{S_{eq}}$ is defined as the following equivalent sectional radius of a wheel:
\begin{eqnarray}
R_{S_{eq}} = \frac{L_{r}}{L}R_{S_{f}} + \frac{L_{f}}{L}R_{S_{r}}. \label{R_S_eq_app}
\end{eqnarray}
$u_{v}$ represents the lateral acceleration term generated by the vehicle' s propulsion.
\begin{eqnarray}
u_{v} = \dot{V}_{oy} + V_{ox}\omega_{z}. \label{u_v_org}
\end{eqnarray}
$V_{oy}$ and $\omega_{z}$ denote the translational velocity in the $y$-direction and the angular velocity about the $z$-axis of the origin of the $x$-$y$-$z$ coordinate system, respectively. 
Assuming that the following geometric relationships hold while neglecting the effects of the tire slip, $V_{oy}$ and $\omega_{z}$ can be expressed as follows: 
\begin{eqnarray}
V_{oy} &=& \frac{L_{r}V_{ox}\cos\theta_{cf}}{L}\delta_{f} + \frac{L_{f}V_{ox}\cos\theta_{cr}}{L}\delta_{r},  \label{V_oy_app}\\
\omega_{z} &=& \frac{V_{ox}\cos\theta_{cf}}{L}\delta_{f} - \frac{V_{ox}\cos\theta_{cr}}{L}\delta_{r}. \label{omega_z_app}
\end{eqnarray}
Here, $V_{ox}$ denotes the translational velocity in the $x$-direction, which is calculated from the angular velocity of either the front or rear wheel.
$u_{v_{0}}$ in (\ref{u}) is the lateral acceleration term, which occurs due to the steering of the front and rear wheels. It does not depend on the vehicle speed and can be expressed as follows:
\begin{eqnarray}
u_{v_{0}} = C_{\delta_{f}}\delta_{f} + C_{\delta_{r}}\delta_{r}. \label{u_v0}
\end{eqnarray}
Here, $C_{\delta_{i}}$ is expressed as follows:
\footnotesize
\begin{eqnarray}
& C_{\delta_{i}} = \frac{gL_{\bar{i}}\left( L_{oi}  -  R_{i}\sin\theta_{ci} \right)}{m_{1}h'L} \left( m_{2} +  \frac{mR_{S_{eq}}}{h'} + \frac{mR_{S_{i}}\sin\theta_{ci}}{L_{oi}  - R_{i}\sin\theta_{ci}} \right),  \nonumber \\ & \label{c_delta} 
\end{eqnarray}
\normalsize
$\bar{i}$ is an index that is assigned $r$ in the case of $i=f$ and $f$ in the case of $i=r$.
Since $L_{or}$ is a nonlinear function of $\delta_{r}$, $C_{\delta_{r}}$ is also a
nonlinear function of $\delta_{r}$. \\
Note that, from (\ref{u_v_org}) and (\ref{V_oy_app}), $u_{v}$ also includes the steering angular velocities of the front and rear wheels.
\subsection{Divergent component of motion feedback control}
The DCM (e.g., \cite{takenaka_1} -- \cite{englsberger_2}) is a quantity that emerges when a biped robot is modeled as an LIP and its dynamics are decomposed via eigenvalue analysis. It represents the unstable mode that grows over time. For biped robots, there exists a constraint that the zero moment point (ZMP) must remain within the support polygon. Sugihara \cite{sugihara} showed that for an LIP system with such input constraints, feedback control of the DCM is the best stabilizing regulator. Furthermore, Sumioka \cite{sumioka} demonstrated that a motorcycle can be modeled as a constrained LIP and that feedback control of the DCM enables cooperative balance assistance with the rider. In this study, we adopt DCM feedback control from the perspective of optimal stabilization.

To implement DCM feedback control, $u$ must satisfy the following \cite{sumioka}:
\begin{eqnarray}
u_{DCM} = K_{\xi}\left( P_{1_{y}} -  P_{1_{y_{d}}} \right) + \frac{K_{\xi}}{\omega} \left( \dot{P}_{1_{y}} - \dot{P}_{1_{y_{d}}} \right), \label{u_FB_d}
\end{eqnarray}
where $K_{\xi}$ is the feedback gain. In this study, we set $K_{\xi} = 2\omega^2$ so that the poles become multiple roots. 
$P_{1_{y_{d}}}$ and $\dot{P}_{1_{y_{d}}}$ are the desired values of $P_{1_{y}}$ and $\dot{P}_{1_{y}}$, respectively.

\section{METHOD}
This section describes the hardware configuration and remote operation method. Furthermore, it explains the proposed approach for incorporating road surface friction into the E2PMD-LIP model, as well as the front- and rear-wheel steering control method that accounts for this factor.
\subsection{Configuration for remote operation}
\begin{figure*}
\centering
\resizebox*{15cm}{!}{\includegraphics{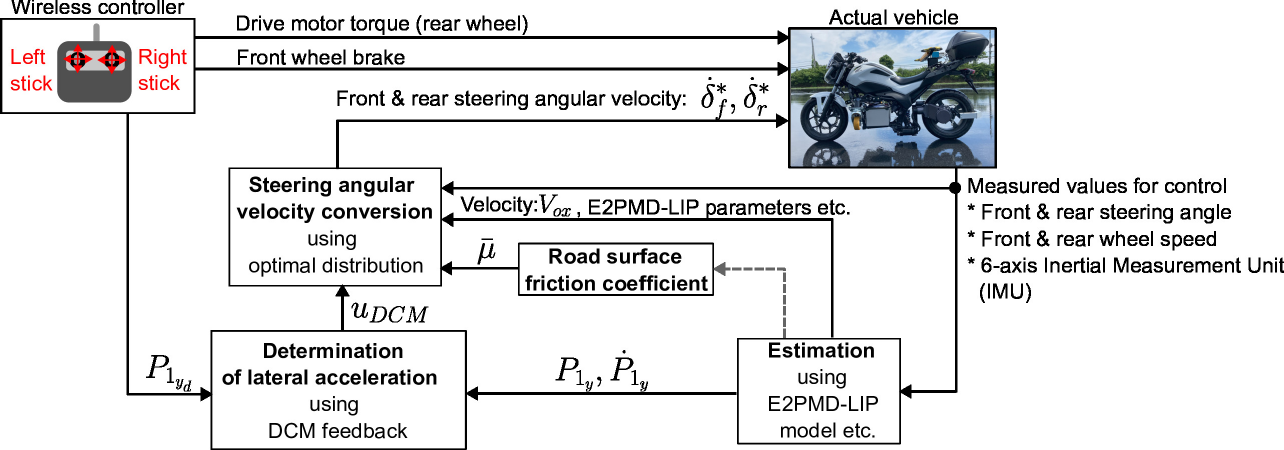}}\hspace{5pt}
\caption{Diagram of overall control scheme. The left stick of the wireless controller was configured such that, when fully tilted to the left, the command becomes $P_{1yd} = |P_{1yd_{\lim}}|$, and when fully tilted to the right, it becomes $P_{1yd} = -|P_{1yd_{\lim}}|$. The lateral analog values of the left stick were linearly interpolated between these limits and transmitted to the onboard controller.
Similarly, when the left stick is pushed fully downward, the vertical analog values of the left stick were linearly interpolated so that a larger front‑brake‑lever command is generated, and this signal was transmitted to the onboard controller.
Furthermore, the vertical analog values of the right stick were configured such that upward input corresponds to a positive motor torque and downward input corresponds to a negative motor torque, and this command was also transmitted to the onboard controller.} \label{fig:control_scheme}
\end{figure*}

The vehicle shown in Fig.~\ref{fig:riding_assist_vehicle} is configured to enable remote operation, allowing transmission of $P_{1_{y_{d}}}$ via a wireless controller, as illustrated in Fig.~\ref{fig:control_scheme}\footnote{In this study, we set $\dot{P}_{1_{y_{d}}}=0$.}.
The rear-wheel drive torque command value can also be transmitted via the wireless controller. A servo motor is mounted on the front-wheel brake lever, allowing remote adjustment of the lever's position via a wireless controller. 

Although a Global Navigation Satellite System is not used for the controller, it is used to measure the side-slip angle around the center of gravity. 
\subsection{Road friction coefficient compensation}
$u_{v0}$ is the lateral acceleration generated by the movement of the contact point that occurs when steering the front and rear wheels and the movement of the center of gravity based on the law of conservation of angular momentum\cite{sumioka}. This term is thus independent of road friction. 
On the other hand, $u_{v}$ is the lateral acceleration caused by the force acting between the tire and the road surface while driving. This term thus depends on road friction. 
As noted above, (\ref{u_v_org}) was derived under the assumption of no tire slip; therefore, we consider compensating for the slip.

We redefine (\ref{u}) as follows:
\begin{eqnarray}
u = \bar{\mu} u_{v} + u_{v_{0}}, \label{u_mu}
\end{eqnarray}
here, $\bar{\mu}$ is a variable that changes according to the road friction conditions, and it is essentially assigned a value corresponding to the actual road friction coefficient.
By constructing the controller based on a simple model in which a decrease in the road friction coefficient leads to a loss of tire grip and, consequently, a reduction in the achievable lateral acceleration, the controller generates a large control input when slip occurs, thereby preventing a fall.
Although real‑time estimation of $\bar{\mu}$ is desirable in the future, in this study it is treated as known in order to first verify the effectiveness of the proposed method.
When $\theta_{cr} \simeq \pi/2$ rad, it follows from (\ref{V_oy_app}) and (\ref{omega_z_app}) that $u_{v}$ is minimally affected by rear-wheel steering and is predominantly governed by the vehicle speed and the front-wheel steering angle.
In addition, it follows from (\ref{c_delta}) that $u_{v_{0}}$ is predominantly governed by the rear-wheel steering angle.
\par
\begin{figure}
\centering
\resizebox*{9cm}{!}{\includegraphics{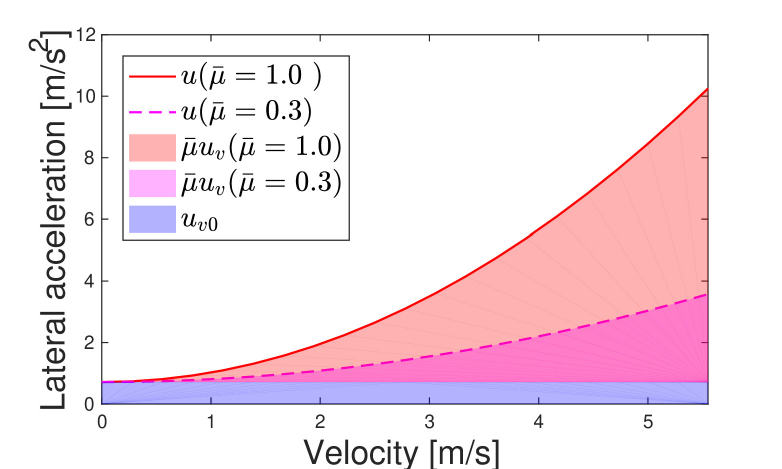}}\hspace{5pt}
\caption{Lateral acceleration generated when steering front and rear wheels at 0.524 rad and 0.175 rad, respectively, with respect to straight-ahead driving state.} \label{fig:u_v}
\end{figure}

Fig.~\ref{fig:u_v} shows $u$, $\bar{\mu} u_{v}$, and $u_{v0}$ that occur when the front and rear wheels are steered at 0.524 rad and 0.175 rad, respectively, with respect to the straight-ahead driving state. 
At extremely low speeds, $u_{v_{0}}$ is dominant; however, as the vehicle speed increases, $\bar{\mu}u_{v}$ becomes dominant. 
In other words, rear-wheel steering is dominant at extremely low speeds and front-wheel steering is dominant at normal speeds. 
Furthermore, when $\bar{\mu}$ decreases from 1.0 to 0.3, the ratio of $\bar{\mu}u_{v}$ to $u_{v_{0}}$ changes. This figure shows that the wheel that should be assisted depends on the vehicle speed and road friction conditions.

\subsection{Optimal distribution of front- and rear-wheel steering}
If the inputs of this system are $\dot{\delta}_{f}$ and $\dot{\delta}_{r}$ and the outputs are $P_{1_{y}}$ and $\dot{P}_{1_{y}}$, the system becomes a two-input, two-output system and can thus be stabilized using a linear quadratic regulator. 
However, constructing the controller using a linear quadratic regulator would make it difficult to maintain the DCM feedback control configuration, which is the best regulator in terms of stabilizing the LIP with input constraints\cite{sugihara}.  
Therefore, in this study, we propose a method that can calculate the optimal assist distribution to the front and rear wheels while maintaining the DCM feedback configuration and taking into account the vehicle speed and road friction conditions.
\par
From (\ref{V_oy_app}), $\dot{V}_{oy}$ can be calculated as follows:
\begin{eqnarray}
\dot{V}_{oy} = \frac{\partial V_{oy}}{\partial \delta_{f}}\dot{\delta}_{f} + \frac{\partial V_{oy}}{\partial \delta_{r}}\dot{\delta}_{r} + \frac{\partial V_{oy}}{\partial V_{ox}}\dot{V}_{ox}. \label{V_oy_dot}
\end{eqnarray}
Therefore, using (\ref{u_v_org}) and (\ref{V_oy_dot}), (\ref{u_mu}) can be rewritten as follows:
\begin{eqnarray}
u &=& \bar{\mu} \frac{\partial V_{oy}}{\partial \delta_{f}}\dot{\delta}_{f} + \bar{\mu} \frac{\partial V_{oy}}{\partial \delta_{r}}\dot{\delta}_{r} + \bar{\mu} \frac{\partial V_{oy}}{\partial V_{ox}}\dot{V}_{ox} \nonumber \\
&+& \bar{\mu} V_{ox}\omega_{z} + u_{v_{0}}. \label{u_detail}
\end{eqnarray}
Here, we collectively denote the fourth and fifth terms on the right-hand side of (\ref{u_detail}) as $u'$:
\begin{eqnarray}
u' = \bar{\mu} V_{ox}\omega_{z} + u_{v_{0}}. \label{u_dash}
\end{eqnarray}
Since $u'$ is a nonlinear function of $\delta_{r}$, we consider expressing it as a function of $\dot{\delta}_{f}$ and $\dot{\delta}_{r}$ using a Taylor approximation around the point in the previous sampling period as follows:
\begin{eqnarray}
u' &\simeq& u'_{p} + \frac{\partial u'}{\partial \delta_{f}}\left( \delta_{f} - \delta_{f_{p}} \right) + \frac{\partial u'}{\partial \delta_{r}}\left( \delta_{r} - \delta_{r_{p}} \right) \nonumber \\
&=& u'_{p} + \frac{\partial u'}{\partial \delta_{f}}\Delta t \dot{\delta}_{f} + \frac{\partial u'}{\partial \delta_{r}}\Delta t \dot{\delta}_{r}. \label{u_dash_app}
\end{eqnarray}
Here, $u'_{p}$, $\delta_{f_{p}}$, and $\delta_{r_{p}}$ are the values of $u'$, $\delta_{f}$, and $\delta_{r}$ from the previous sampling period, respectively. $\Delta t$ is the sampling period. In this study, we set $\Delta t = 0.001$. From (\ref{u_detail}), (\ref{u_dash}), and (\ref{u_dash_app}), $u$ can be approximated as follows:
\begin{eqnarray}
u &\simeq& \left( \bar{\mu} \frac{\partial V_{oy}}{\partial \delta_{f}} + \frac{\partial u'}{\partial \delta_{f}}\Delta t \right)\dot{\delta}_{f} + \left( \bar{\mu} \frac{\partial V_{oy}}{\partial \delta_{r}} + \frac{\partial u'}{\partial \delta_{r}}\Delta t \right)\dot{\delta}_{r} \nonumber \\
&+& \bar{\mu} \frac{\partial V_{oy}}{\partial V_{ox}}\dot{V}_{ox} + u'_{p}. \label{u_detail2}
\end{eqnarray}
Consider solving the following optimization problem with the constraint that the right-hand side of (\ref{u_detail2}) is equal to the right-hand side of (\ref{u_FB_d}). 
\begin{eqnarray}
&{\rm arg}& \min_{\dot{\delta}_{f},\dot{\delta}_{r}} \left( w_{f}\dot{\delta}_{f}^{2} + w_{r}\dot{\delta}_{r}^{2} \right) \label{opt_fr1} \\
&{\mathrm{s.t.}}& \left( \bar{\mu}\frac{\partial V_{oy}}{\partial \delta_{f}} + \frac{\partial u'}{\partial \delta_{f}}\Delta t \right)\dot{\delta}_{f} + \left( \bar{\mu}\frac{\partial V_{oy}}{\partial \delta_{r}} + \frac{\partial u'}{\partial \delta_{r}}\Delta t \right)\dot{\delta}_{r} \nonumber \\
&& = u_{DCM} - u'_{p} - \bar{\mu}\frac{\partial V_{oy}}{\partial V_{ox}}\dot{V}_{ox},  \label{opt_fr2} \\
&& w_{f} > 0, w_{r} > 0.  \label{opt_fr3}
\end{eqnarray}
Here, $w_{f}$ and $w_{r}$ are weight variables. Their default values are $w_{f}=1$ and $w_{r}=1$. The weight on the side with a larger tire friction circle limit margin can be increased if necessary. 
Since the above optimization problem is a linear quadratic programming problem, the optimal solutions $\dot{\delta}_{f}^{*}$ and $\dot{\delta}_{r}^{*}$ can be found without iterative calculations as follows: 
\footnotesize
\begin{eqnarray}
\dot{\delta}_{f}^{*} \! \! \! \! &=&\! \! \! \!  \frac{w_{r}\left( u_{DCM} - u'_{p} - \bar{\mu}\frac{\partial V_{oy}}{\partial V_{ox}}\dot{V}_{ox} \right) \left( \bar{\mu}\frac{\partial V_{oy}}{\partial \delta_{f}} + \frac{\partial u'}{\partial \delta_{f}} \Delta t\right)}{w_{f}\left( \bar{\mu}\frac{\partial V_{oy}}{\partial \delta_{r}} + \frac{\partial u'}{\partial \delta_{r}}\Delta t \right)^2 + w_{r}\left( \bar{\mu}\frac{\partial V_{oy}}{\partial \delta_{f}} + \frac{\partial u'}{\partial \delta_{f}}\Delta t \right)^2} , \label{opt_delta_f_dot} \\
\dot{\delta}_{r}^{*} \! \! \! \! &=&\! \! \! \!  \frac{w_{f}\left( u_{DCM} - u'_{p} - \bar{\mu}\frac{\partial V_{oy}}{\partial V_{ox}}\dot{V}_{ox} \right) \left( \bar{\mu}\frac{\partial V_{oy}}{\partial \delta_{r}} + \frac{\partial u'}{\partial \delta_{r}} \Delta t\right)}{w_{f}\left( \bar{\mu}\frac{\partial V_{oy}}{\partial \delta_{r}} + \frac{\partial u'}{\partial \delta_{r}}\Delta t \right)^2 + w_{r}\left( \bar{\mu}\frac{\partial V_{oy}}{\partial \delta_{f}} + \frac{\partial u'}{\partial \delta_{f}}\Delta t \right)^2}. \label{opt_delta_r_dot}
\end{eqnarray}
\normalsize
The term $ \bar{\mu}(\partial V_{oy}/\partial \delta_{i}) + (\partial u'/\partial \delta_{i})\Delta t$ determines the distribution ratio based on the vehicle speed and road friction conditions. The most efficient input is selected based on the cost function of input minimization. 

In addition, $V_{oy}$ and $u'$ are variables that depend on vehicle parameters such as $\theta_{cr}$. Although the vehicle used in this study is $\theta_{cr} \simeq \pi/2$, the proposed method is also applicable to vehicles with, for example, $\theta_{cr} \simeq 0$. In such cases, the optimal distribution that accounts for the corresponding vehicle parameters is achieved through the term $\bar{\mu}(\partial V_{oy}/\partial \delta_{i}) + (\partial u'/\partial \delta_{i})\Delta t$.

\section{RESULTS}
\begin{figure*}
\centering
\resizebox*{12cm}{!}{\includegraphics{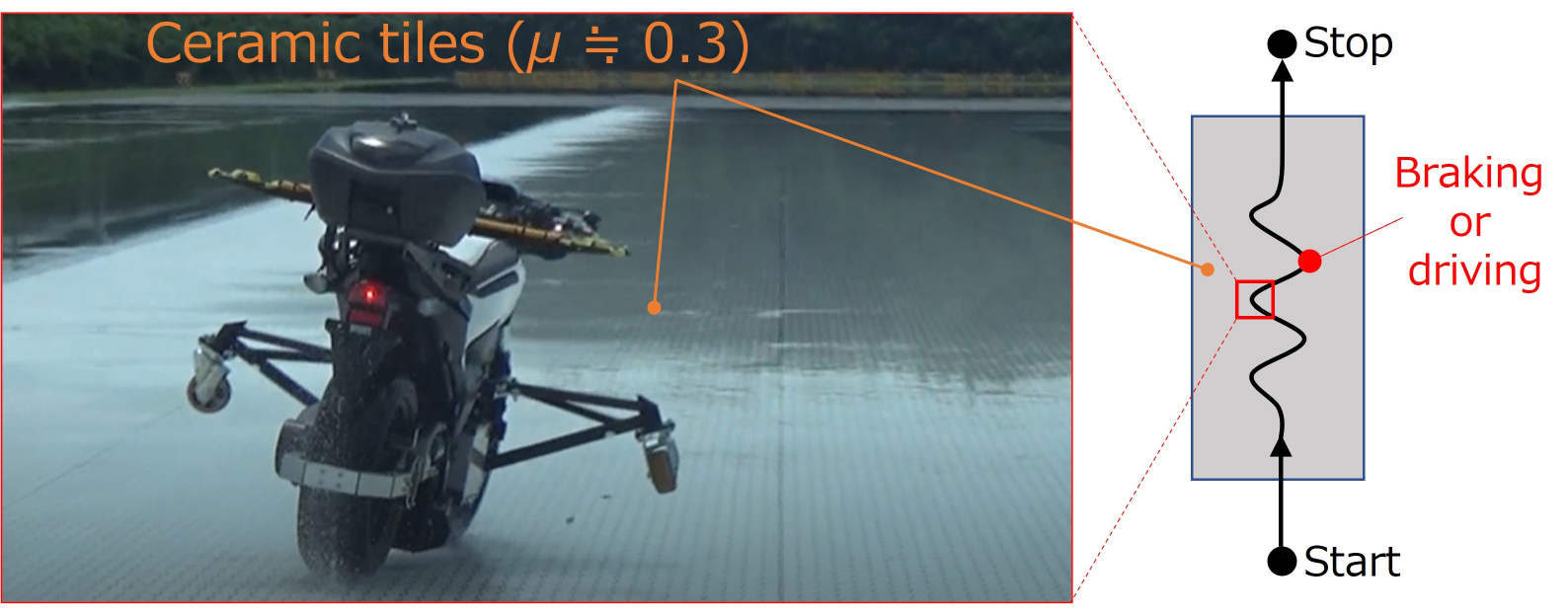}}\hspace{5pt}
\caption{Snapshot of experiment of low-$\mu$ road driving. A slalom experiment was conducted on wet ceramic tiles. Slip was induced by braking or driving at a turning point of the slalom run.} \label{fig:test_course}
\end{figure*}

\begin{figure*}
\centering
\resizebox*{12cm}{!}{\includegraphics{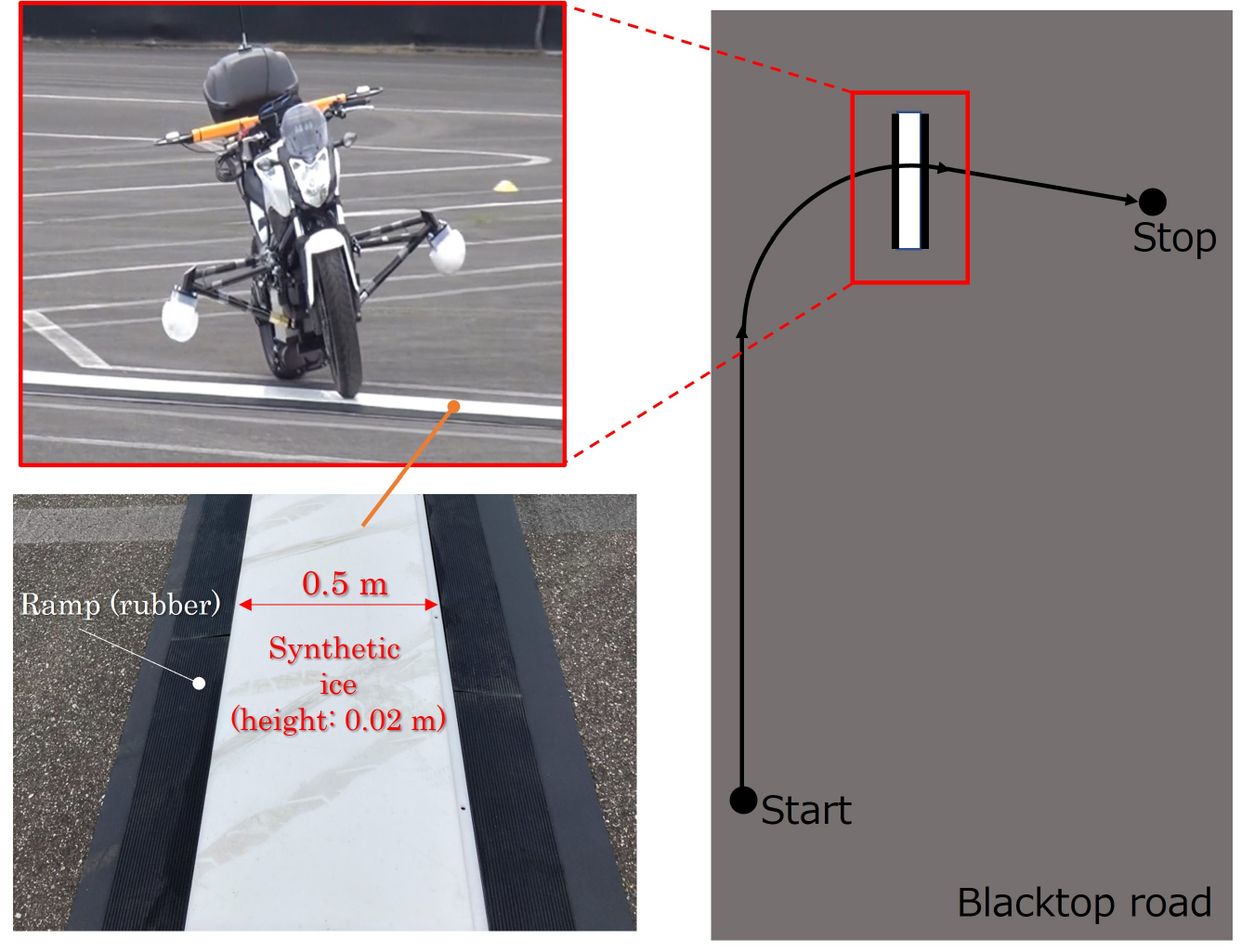}}\hspace{5pt}
\caption{Snapshots of experiment using synthetic ice. The coefficient of friction of the strip of synthetic ice was approximately 0.36.} \label{fig:ICE_rink}
\end{figure*}

\par 
As shown in Fig.~\ref{fig:test_course}, the vehicle was driven in a slalom pattern at a speed of 5.5 m/s on a low-$\mu$ road ($\mu = 0.3$). 
In addition, tests were conducted in which slip was induced by braking the front or rear wheels at points corresponding to the inflection points of the slalom run and in which slip was induced by driving the rear wheel. 
\par 
A typical example of slipping while turning is where the front wheels slip when passing over a manhole in rainy weather. 
If the coefficient of friction between the front wheel and the road surface changes suddenly, the vehicle may fall over due to the body of the vehicle swinging significantly to the outside of the turn after the front wheel regains grip. 
Therefore, we conducted a test in which the vehicle made a turn around a strip of synthetic ice with a width equivalent to the diameter of a manhole at a speed of 5.5 m/s while braking the front wheels, as shown in Fig.~\ref{fig:ICE_rink}. 
\par 
In the following, we primarily present the results obtained using the proposed method with a fixed value of $\bar{\mu} = 0.3$, along with those obtained using a conventional method. 
The $\phi_{b}$ and $\omega_{z}$ values were measured by an inertial measurement unit and the $\beta$ values were measured by a Global Navigation Satellite System. 

\subsection{Conventional method}
We implemented a controller for a riderless two-wheel vehicle \cite{miyagishi}, \cite{auto_bicycle} that determines the front-wheel steering angle based on the deviation between the desired roll angle and the actual roll angle.
\begin{eqnarray}
\delta_{f} = K_{1}\left( \phi_{b} - \phi_{b_{d}} \right) + K_{2}\dot{\phi}_{b}, \label{ARAS_CONV}
\end{eqnarray}
where $K_{1}$ and $K_{2}$ are set as gains determined from the vehicle specifications, as described in previous studies\cite{auto_bicycle}. $\phi_{b_{d}}$ is the desired roll angle specified by the wireless controller.
Since the conventional method (\ref{ARAS_CONV}) was unable to achieve self-balancing when stopped, straight-line driving from a stopped self-supported state to the target vehicle speed of 5.5 m/s was performed using the proposed method and then testing was conducted by switching to the conventional method. 
As a result, the conventional method, which does not account for the road friction conditions or the vehicle dynamics model, generated insufficient front-wheel steering input during slip, leading to falling over in all turning maneuvers on low-friction surfaces.
\subsection{Slalom test results}
\begin{figure}
\centering
\resizebox*{9cm}{!}{\includegraphics{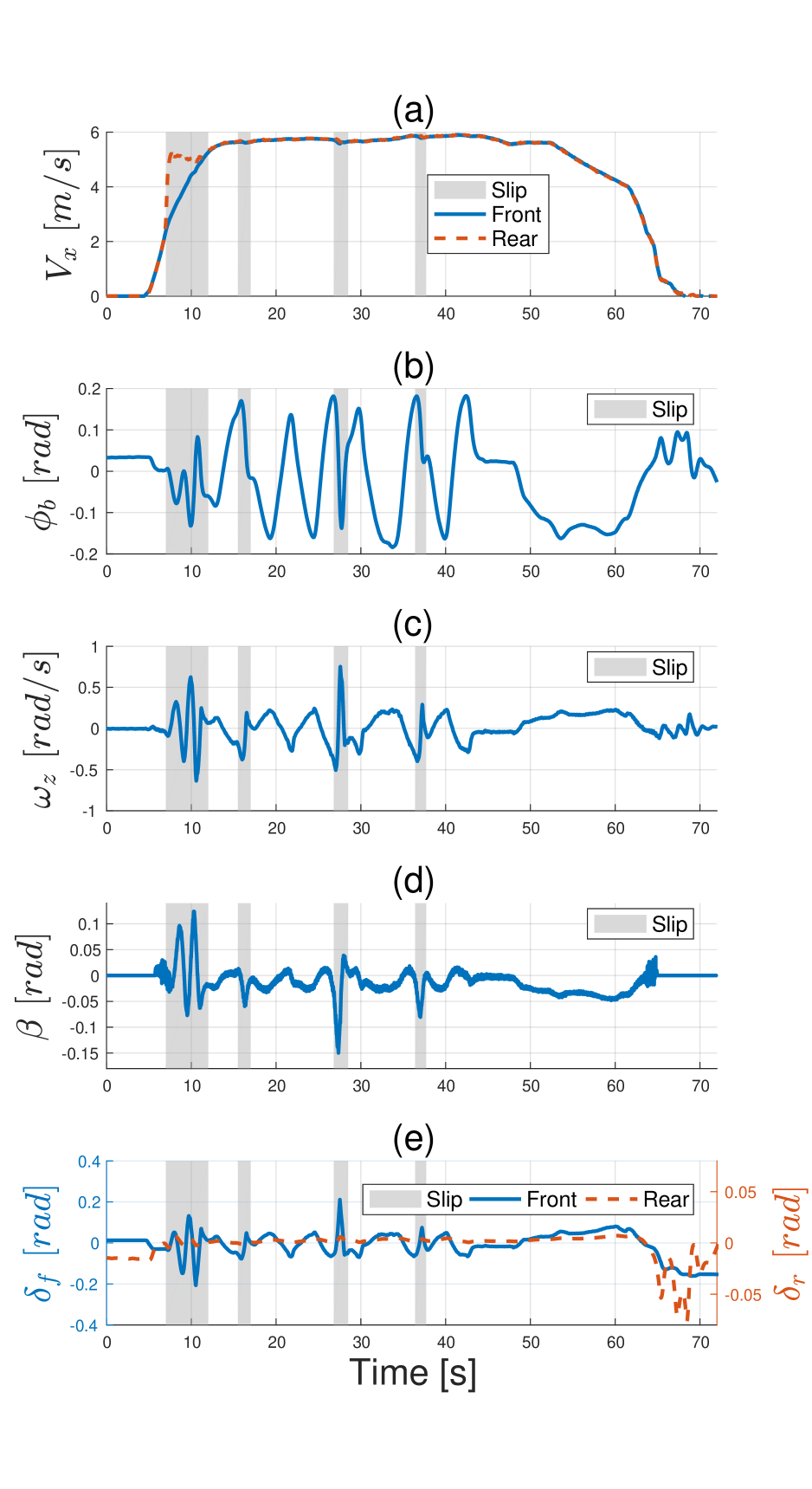}}\hspace{5pt}
\caption{Results of low-$\mu$ road slalom test. (a) Longitudinal velocity. (b) Vehicle body roll angle. (c) Vehicle body yaw rate. (d) Side-slip angle of center of gravity. (e) Front- and rear-wheel steering angles. Slips occurred at around 10, 15, 28, and 37 s.} \label{fig:sla}
\end{figure}

Fig.~\ref{fig:sla} shows the results of driving at a speed of 5.5 m/s on a low-$\mu$ road without braking or driving at the inflection point of the slalom run. 
It can be seen that rear-wheel steering is used when stopped or moving at extremely low speeds. When the vehicle speed increases to a certain level, front-wheel steering is used. 
In the normal driving speed range, the front-wheel steering control plays the dominant role; however, unlike the conventional method, the steering input is computed by taking into account both the road friction conditions and the vehicle states. As a result, the vehicle can avoid falling over.
In addition, at around 10 s, the rear wheel spins during acceleration, causing a large yaw rate and side-slip angle while the vehicle slips; nevertheless, the vehicle is still able to move without falling over. 
It can also be seen that a large yaw rate and a large side-slip angle occur at the inflection points of the slalom run (at around 15, 28, and 37 s) and that the vehicle slips but is still able to move without falling over.

\subsection{Results of front-wheel braking during slalom test}
\begin{figure}
\centering
\resizebox*{9cm}{!}{\includegraphics{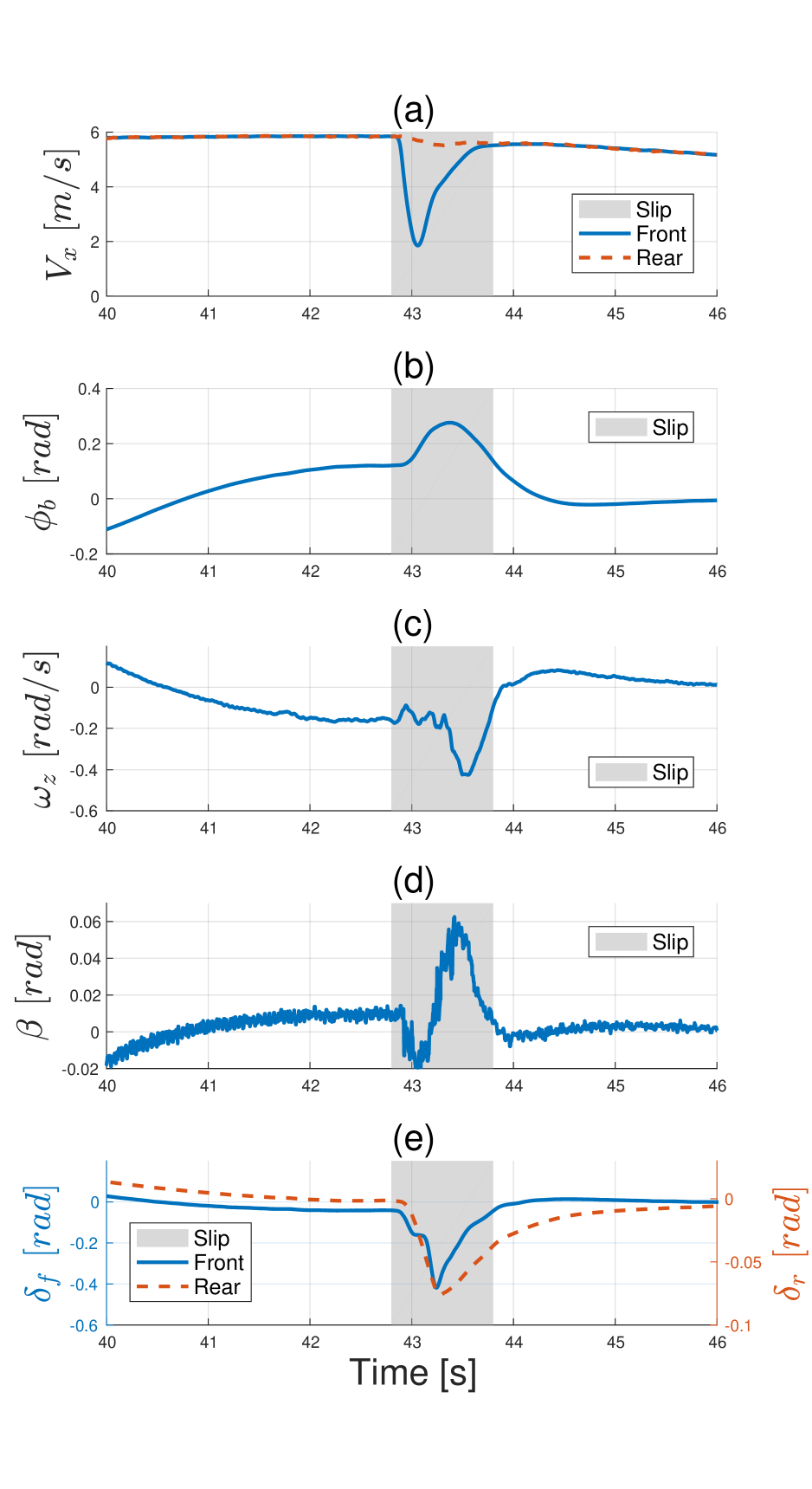}}\hspace{5pt}
\caption{Results of front-wheel braking during low-$\mu$ slalom test. (a) Longitudinal velocity. (b) Vehicle body roll angle. (c) Vehicle body yaw rate. (d) Side-slip angle of center of gravity. (e) Front- and rear-wheel steering angles. Slips occurred at around 43 s.} \label{fig:Fr_BRK}
\end{figure}

The results of applying the front-wheel brakes at an inflection point during the slalom test are shown in Fig.~\ref{fig:Fr_BRK}. 
Figs.~\ref{fig:Fr_BRK}(a) and (d) show that a combined slip state occurs due to a slip ratio of approximately -67$\%$ and a large side-slip angle, but falling over is suppressed. 

\subsection{Results of rear-wheel braking during slalom test}
\begin{figure}
\centering
\resizebox*{9cm}{!}{\includegraphics{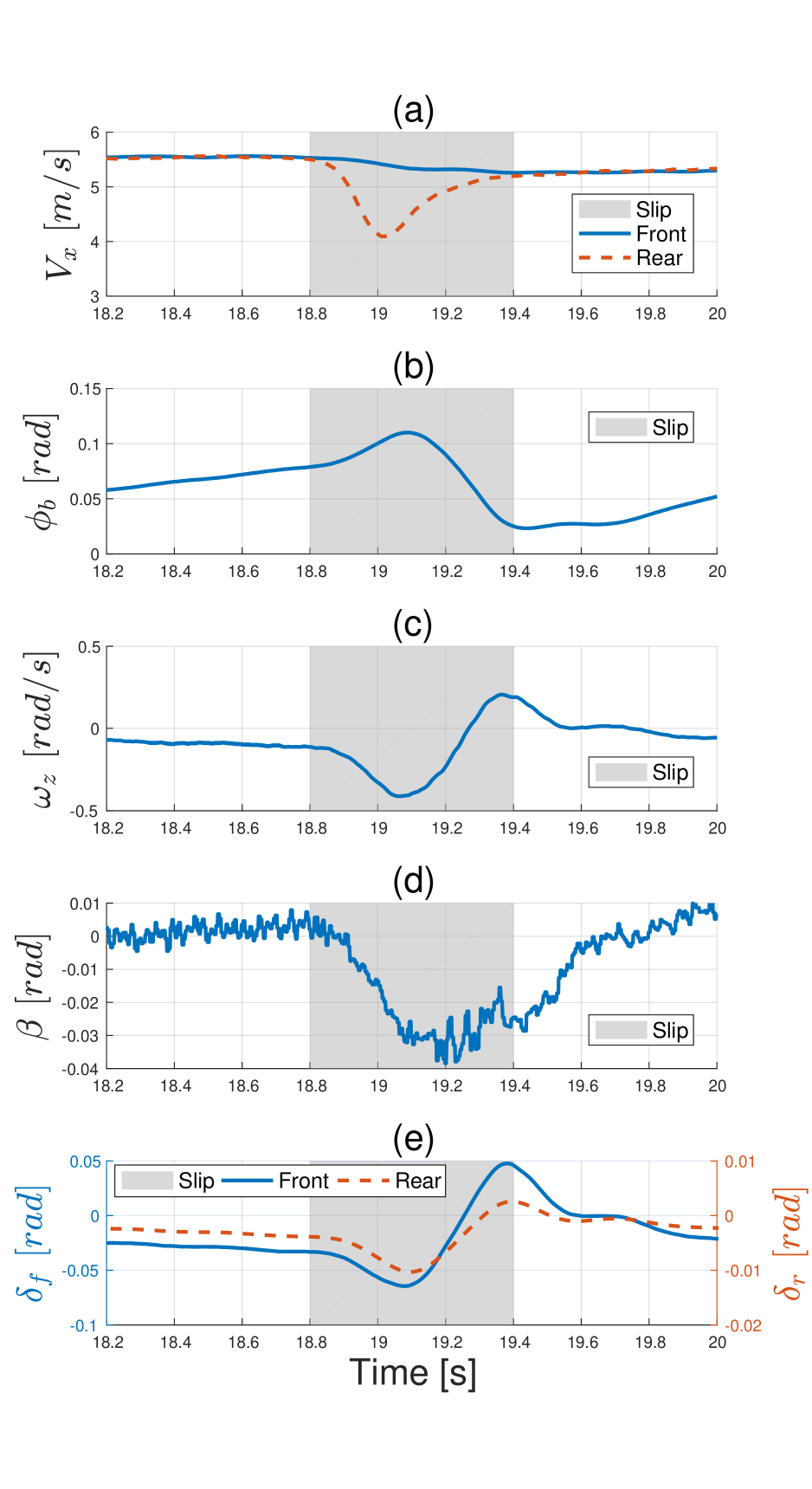}}\hspace{5pt}
\caption{Results of rear-wheel braking during slalom test on low-$\mu$ road. (a) Longitudinal velocity. (b) Vehicle body roll angle. (c) Vehicle body yaw rate. (d) Side-slip angle of center of gravity. (e) Front- and rear-wheel steering angles. Slips occurred at around 19 s.} \label{fig:Rr_BRK}
\end{figure}

The results of applying the rear-wheel brakes at an inflection point during the slalom test are shown in Fig.~\ref{fig:Rr_BRK}. 
Figs.~\ref{fig:Rr_BRK}(a) it can be seen that a longitudinal slip ratio of approximately -24$\%$ is generated. 
In addition, on a typical dry asphalt road surface, when the vehicle travels at the speed shown in Figs.~\ref{fig:Rr_BRK}(a) and performs a right turn while leaning the vehicle by approximately 0.1 rad, $\beta$ takes a small positive value (see Appendix A1). However, Figs.~\ref{fig:Rr_BRK}(d) indicates that lateral slipping occurs with a negative value of $\beta$; nevertheless, it can be observed that falling over is successfully avoided.

\subsection{Results of rear-wheel driving during slalom test}
\begin{figure}
\centering
\resizebox*{9cm}{!}{\includegraphics{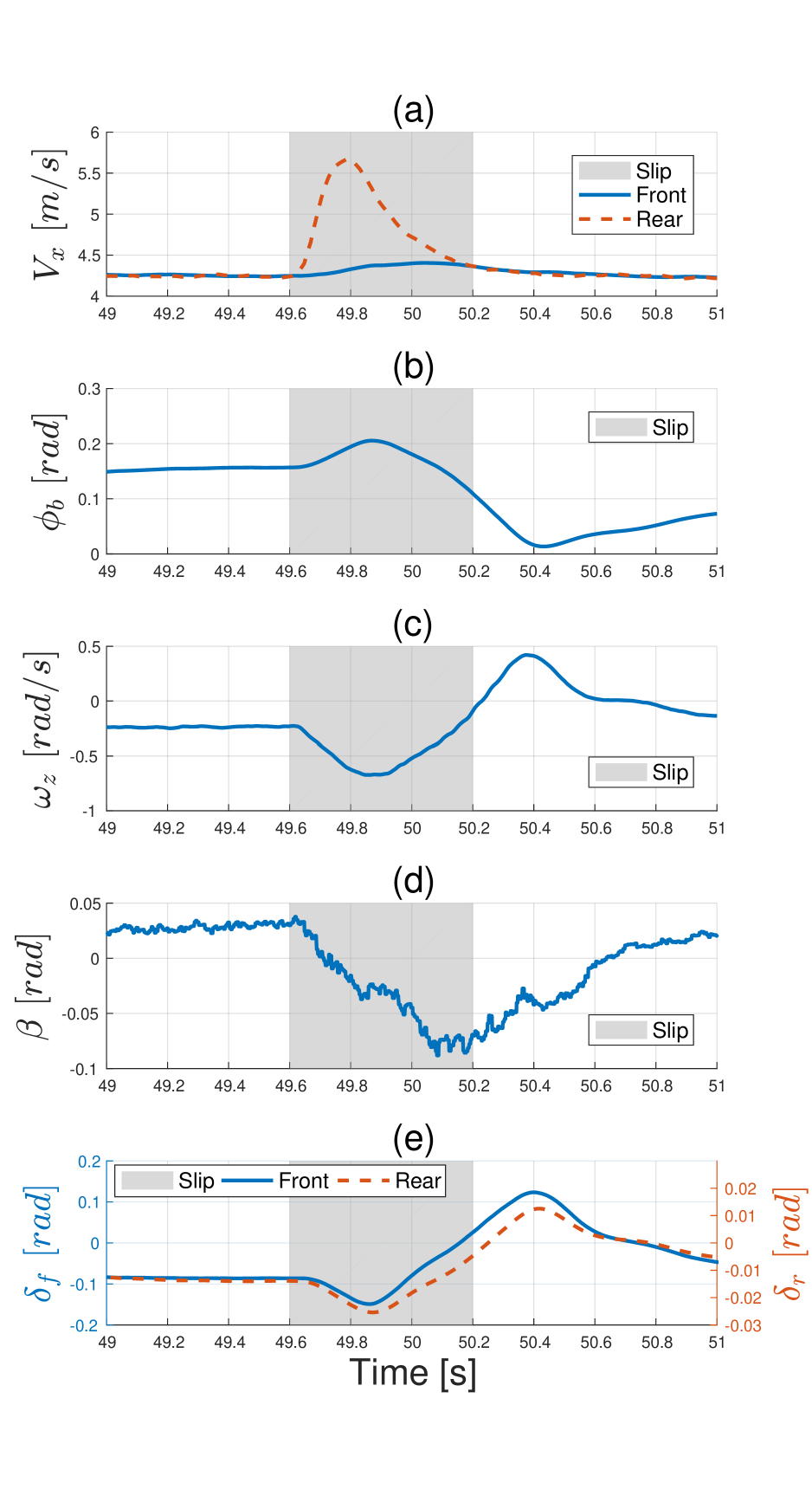}}\hspace{5pt}
\caption{Results of rear-wheel driving during low-$\mu$ slalom test. (a) Longitudinal velocity. (b) Vehicle body roll angle. (c) Vehicle body yaw rate. (d) Side-slip angle of center of gravity. (e) Front- and rear-wheel steering angles. Slips occurred at around 50 s.} \label{fig:Rr_DRV}
\end{figure}

The results obtained with the rear wheel driven at the inflection point during the slalom test are shown in Fig.~\ref{fig:Rr_DRV}. 
Figs.~\ref{fig:Rr_DRV}(a) and (d) show that a combined slip state occurs due to a slip ratio of approximately 31$\%$ and a large side-slip angle, but falling over is suppressed. 

\subsection{Synthetic ice test results}
\begin{figure}
\centering
\resizebox*{9cm}{!}{\includegraphics{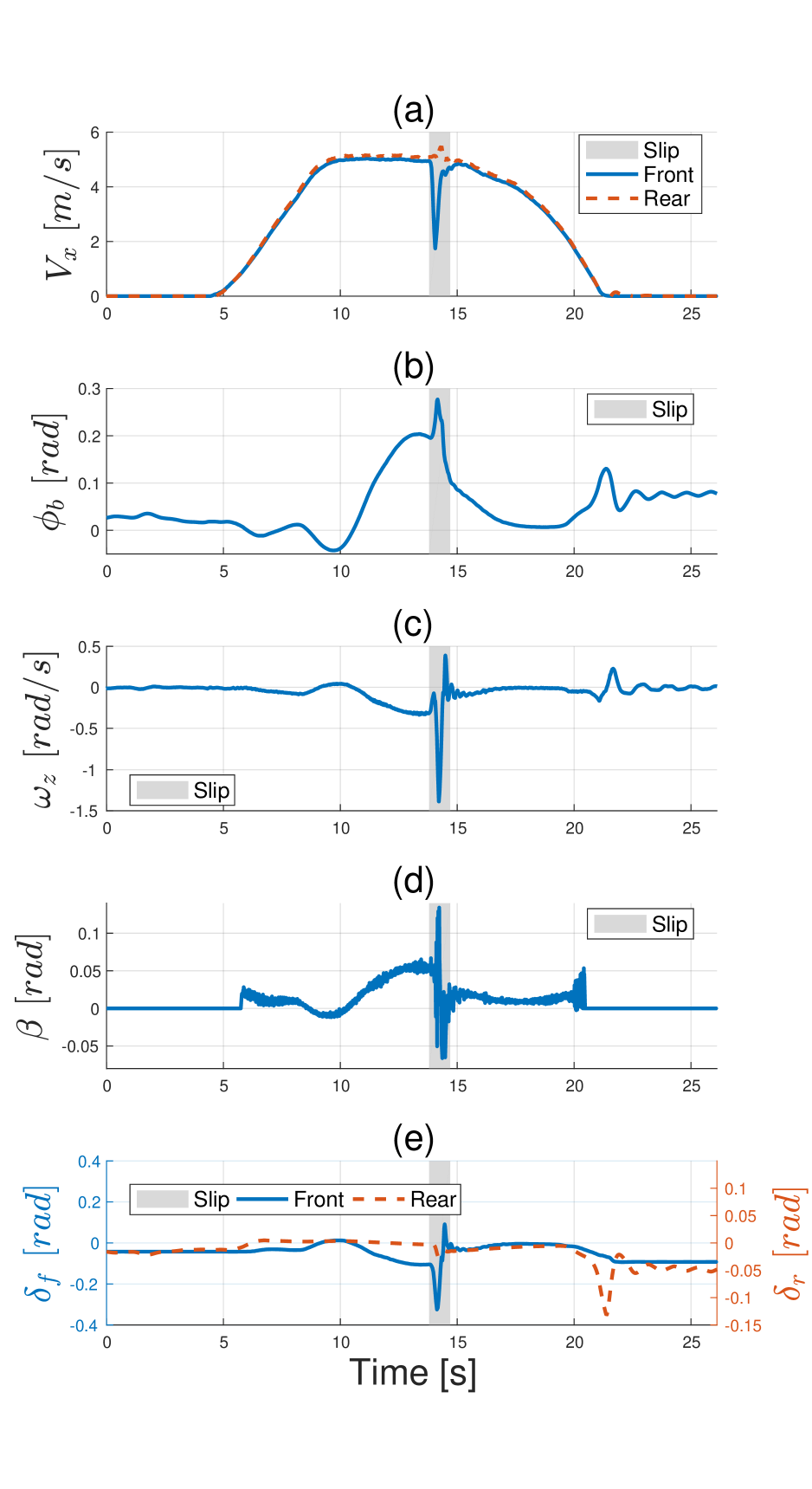}}\hspace{5pt}
\caption{Results of turning on strip of synthetic ice while braking front wheel. (a) Longitudinal velocity. (b) Vehicle body roll angle. (c) Vehicle body yaw rate. (d) Side-slip angle of center of gravity. (e) Front- and rear-wheel steering angles. Slips occurred at around 14 s.} \label{fig:mu_jump}
\end{figure}

The results obtained for turning around on a strip of synthetic ice while braking the front wheel are shown in Fig.~\ref{fig:mu_jump}. 
Figs.~\ref{fig:mu_jump} (a), (c), and (d) show that the slip ratio is approximately -71$\%$, resulting in a combined slip state, and that a side slip occurs along with a large yaw rate; nevertheless, the vehicle does not fall over. 
From Fig.~\ref{fig:mu_jump}(b), it can be seen that although the vehicle roll angle momentarily increases toward the inside of the turn when the side slip occurs, the proposed method prevents rollover by suppressing the subsequent roll motion toward the outside of the turn after the slip period ends.

\section{DISCUSSION}
\par 
When $\bar{\mu}$ was fixed at 1, the vehicle occasionally fell during the low-$\mu$ slalom test. In the synthetic ice test, although falling was avoided, an overshoot in the roll angle toward the high side was observed. These results indicate that road friction coefficient compensation is essential in the proposed method. Furthermore, when the vehicle speed was below 1.4 m/s, slip-induced falling was prevented using only the rear-wheel swing assist. 
However, because $\theta_{cr} \simeq \pi/2$ rad, the lateral acceleration achievable using rear-wheel swing control alone is limited, as shown in Fig.~\ref{fig:u_v} and (\ref{V_oy_app})and (\ref{omega_z_app}). At vehicle speeds exceeding 1.4 m/s, the resulting motor load was too high to maintain prolonged operation, and the motor eventually shut down due to overcurrent protection.
 Nevertheless, at $\bar{\mu}$ = 0.3, the level of assist intervention increased, which may cause discomfort to the rider during driving. 
This is because an increase in control input at $\bar{\mu} = 0.3$ can, as indicated in studies such as \cite{sumioka}, make it more difficult for the rider to lean the motorcycle as intended when initiating a turn. Moreover, even in the higher‑speed operating range considered in this study, a control strategy that preserves rideability while providing fall‑prevention assistance only when a slip is likely to occur will be necessary in the future.
Therefore, future work should focus on establishing a reliable road friction coefficient estimation method and validating the effectiveness of the proposed method under rider-operated conditions. 
\par 
The proposed method suppresses falling over by controlling the steering of the front and rear wheels, so there is a limit to the extent to which the vehicle can suppress falling over if the tire friction circle limit of the front and rear wheels is exceeded. 
For example, if strong braking is applied to both the front and rear wheels simultaneously while cornering on a low-$\mu$ road, the front and rear wheels will enter a combined slip state, causing the vehicle to fall over. 
Therefore, a method for suppressing falling over when strong braking is applied to both the front and rear wheels is a future challenge.
\par In addition, for safety reasons, the upper limit vehicle speed in this experiment was set to 5.5 m/s. Future work should confirm the effectiveness of the proposed method at vehicle speeds higher than 5.5 m/s.

\section{CONCLUSION}
In this study, we proposed a method for preventing the slipping and falling over of a motorcycle while turning and verified it using a vehicle that can steer the front and rear wheels. 
The proposed method optimally distributes the steering assist to the front and rear wheels according to the vehicle speed and road friction coefficient while maintaining DCM feedback control, which is the best regulator for stabilizing an LIP with input constraints. We also showed that the proposed method can be used to make turns while braking or driving at a vehicle speed of 5.5 m/s on a slippery road surface.

\appendix 
\section{Experimental Results on a Dry Road Surface}
\begin{figure*}
\centering
\resizebox*{19cm}{!}{\includegraphics{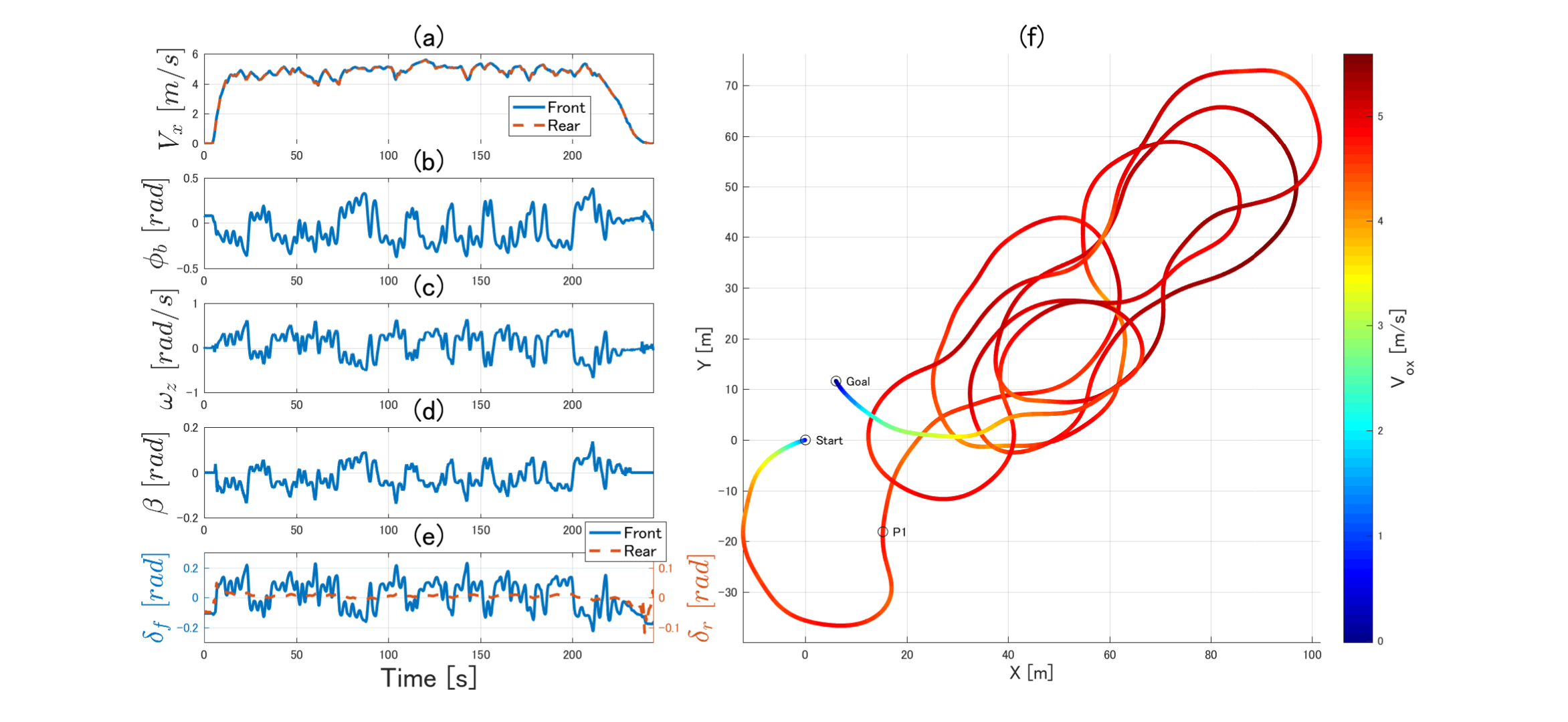}}\hspace{5pt}
\caption{Results of remotely operated driving in an open area without a predefined course on a dry asphalt road surface. (a) Longitudinal velocity. (b) Vehicle body roll angle. (c) Vehicle body yaw rate. (d) Side-slip angle of center of gravity. (e) Front- and rear-wheel steering angles. (f) Trajectory of center of gravity. $\bar{\mu}=1$ was set.} \label{fig:DRY_TEST}
\end{figure*}

\begin{figure}
\centering
\resizebox*{9cm}{!}{\includegraphics{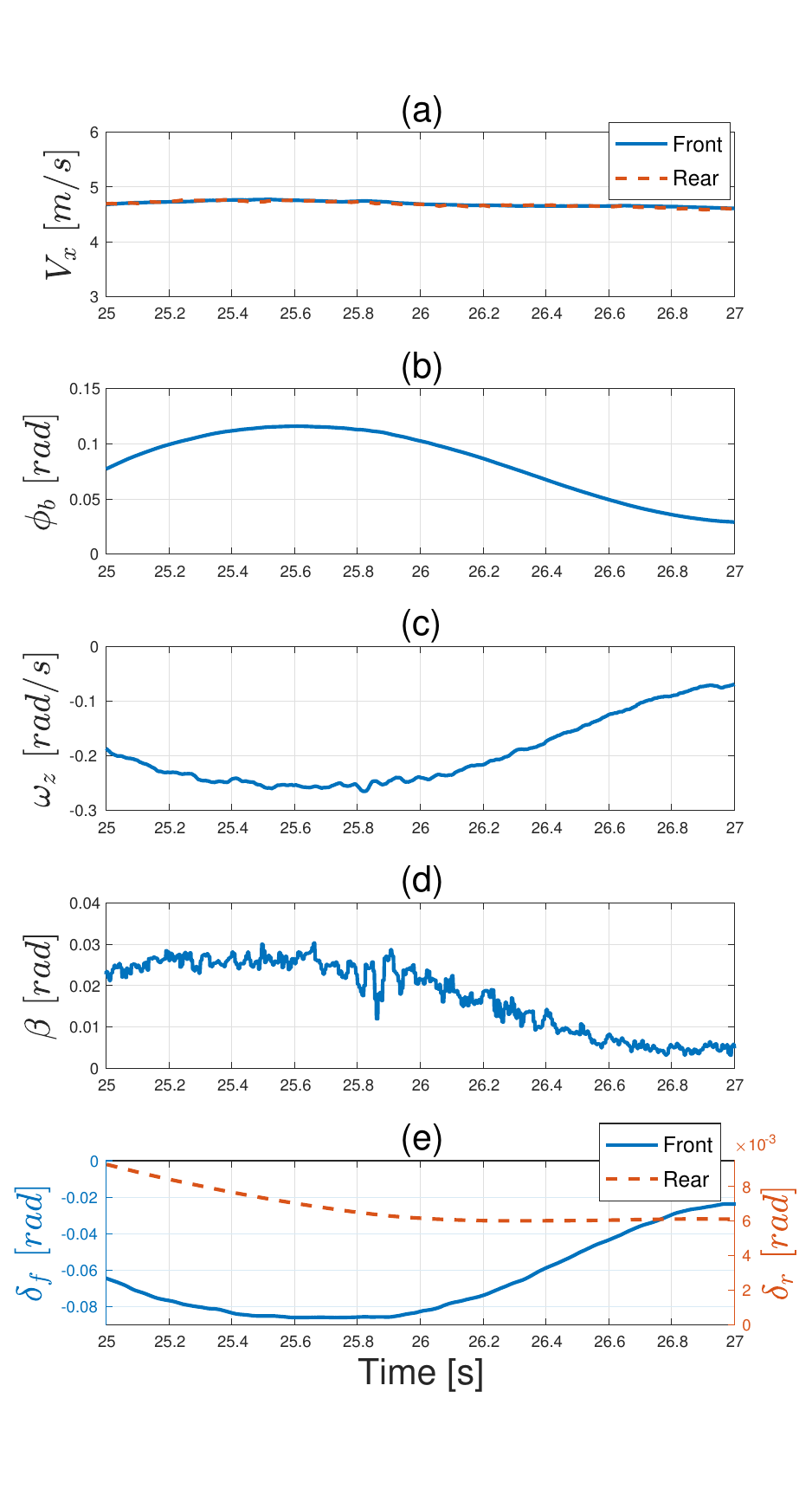}}\hspace{5pt}
\caption{Results extracted around 26 s in Fig.~\ref{fig:DRY_TEST}(around position $P1$ in Fig.~\ref{fig:DRY_TEST}(f)). During right turning with a vehicle lean angle of approximately 0.1 rad at a speed of 4.9 m/s, a small positive $\beta$ was observed.} \label{fig:DRY_P1}
\end{figure}

Fig.~\ref{fig:DRY_TEST} shows the results of remotely operated driving in an open area without a predefined course, conducted with the $\bar{\mu}$ set to 1 on a dry asphalt road surface at a vehicle speed of 5.5 m/s.
To compare with Fig.~\ref{fig:Rr_BRK}, the segment around 26 s in Fig.~\ref{fig:DRY_TEST}, corresponding to $P1$ in Fig.~\ref{fig:DRY_TEST}(f), is shown in Fig.~\ref{fig:DRY_P1}.


\end{document}